\documentclass[10pt,twocolumn]{article} 
\usepackage{simpleConference}
\usepackage{graphicx}
\usepackage{time}
\usepackage{amssymb}
\usepackage{url,hyperref}
\usepackage{cite}
\usepackage{algorithmic}
\usepackage{wrapfig}
\usepackage{amsmath}
\usepackage{amssymb}
\usepackage{booktabs}
\usepackage{multirow}
\usepackage{multicol}
\usepackage{kotex}

\begin{document}

\title{Keypoint based Sign Language Translation without Glosses}

\author{\textbf{Youngmin Kim} \\
\small
Incheon National University \\
\small
Incheon, Republic of Korea \\
\small
winston121497@gmail.com
\and
\textbf{Minji Kwak}\\
\small
Chung-Ang University \\
\small
Seoul, Republic of Korea\\
\small
minz@cau.ac.kr
\and
\textbf{Dain Lee}\\
\small
Ewha Womans University \\
\small
Seoul, Republic of Korea \\
\small
na05128@ewhain.net
\and
\textbf{Yeongeun Kim}\\
\small
Konkuk University \\
\small
Seoul, Republic of Korea \\
\small
zena0101@naver.com
\and
\textbf{Hyeongboo Baek}\\
\small
Incheon National University \\
\small
Incheon, Republic of Korea \\
\small
hbbaek359@gmail.com
}

\maketitle
\thispagestyle{empty}

\begin{abstract}
Sign Language Translation (SLT) is a task that has not been studied relatively much compared to the study of Sign Language Recognition (SLR). 
However, the SLR is a study that recognizes the unique grammar of sign language, which is different from the spoken language and has a problem that non-disabled people cannot easily interpret.
So, we're going to solve the problem of translating directly spoken language in sign language video. 
To this end, we propose a new keypoint normalization method for performing translation based on the skeleton point of the signer and robustly normalizing these points in sign language translation. 
It contributed to performance improvement by a customized normalization method depending on the body parts. 
In addition, we propose a stochastic frame selection method that enables frame augmentation and sampling at the same time. 
Finally, it is translated into the spoken language through an Attention-based translation model. 
Our method can be applied to various datasets in a way that can be applied to datasets without glosses. 
In addition, quantitative experimental evaluation proved the excellence of our method.
\end{abstract}

\section{Introduction}
As the value of equal society has emerged, there is a social atmosphere in which various people should be respected and given equal opportunities.
However, deaf people have difficulty communicating with non-disabled people not only in their daily lives but also in situations where they get the information they need to acquire \cite{one}.
\begin{figure}[h]
    \centering
   \includegraphics[width=1\columnwidth]{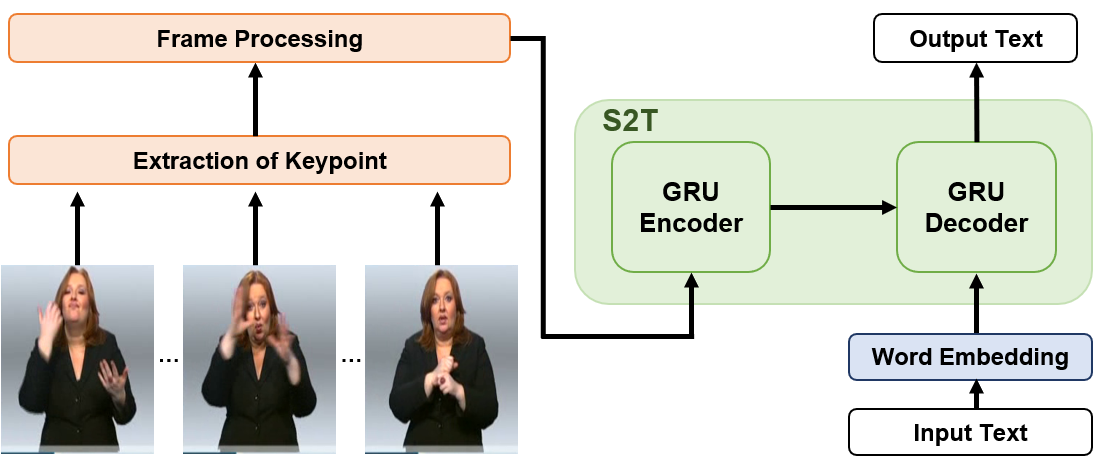}
    \caption{\small \textbf{An overview of Our Approach.} S2T means sign language to text model.}
    \label{fig-label1}
\end{figure}
In order to solve this problem, research on sign language translation has long been conducted to help deaf people live their daily lives using computers \cite{six}. 
These studies are intended to translate sign language videos into spoken language and aim to facilitate smooth communication between the deaf and non-disabled \cite{seven}. 
However, in the field of computer vision, the focus is on Continuous Sign Language Recognition (CSLR), which recognizes successive glosses rather than Sign Language Translation (SLT) \cite{thirteen} which converts sign language video directly into spoken language. 
CSLR is a task that recognizes the grammar of sign language itself, and it is difficult to provide meaningful interpretation only with CSLR because the grammar of sign language and the grammar of the spoken language are different.

Therefore, we have studied a method for translating directly from video to spoken language rather than a translation based on the grammar of sign language, which is possible even in the absence of gloss.
So we try to achieve SLT using Neural Machine Translation (NMT) based methods.
In this paper, we use the GRU \cite{thirtyfour} based Seq2Seq\cite{thirtyfive} model applying the Attention of Bahdanau \textit{et al} \cite{thirtyseven}.

Sign language is a visual language as the main language for communication for the deaf. And sign language, which is a visual language, uses many complementary channels to convey information \cite{eight}.
These include not only hand movements but also nonverbal elements such as the signer's facial expressions, mouth, and upper body movements, which play an important role in communication \cite{nine}.
Therefore, this study proposes a sign language translation method based on the movement of the signer's body.
We perform keypoint-based SLT to more precisely determine the movement of the signer's body.

The frame processing is essential to perform the signer's keypoint-based SLT.
The keypoint shows very different values according to the angle and position of the signer in the frame.
To overcome and generalize the differences accordingly, the normalization method should be applied to the keypoint vector. 
In addition, a keypoint vector of a fixed length is required to perform SLT using the NMT model, and the existing keypoint-based sign language translation models \cite{three,twelve,fourteen,thirty} go through a process of fixing the length of the sign language image through the frame sampling method.
In this process, if the length of the video is short, the possibility that the key frames will not be included increases, which leads to the loss of information on the video.
Conversely, when the length is set to be very large there is a disadvantage that the memory usage is very large while including a lot of frames.

To solve these problems, we propose a simple but effective method.
First, we proposed a normalization method using the distance between each keypoint.
Our normalization method proposes a stronger normalization method by varying normalization depending on the body part.
We called it "Customized Normalization."

Second, in the process of adjusting the length of the sign language video, we propose a method of selecting the frame based on probability. 
Accordingly, it is possible to lower the priority of the less important part of the sign language video and increase the priority of the key frame.
In addition, a method corresponding to the length of a dynamic video frame is proposed by simultaneously using not only the augmentation but also the sampling method depending on the video length.
We called it "Stochastic Augmentation and Skip Sampling(SASS)".

In this paper, we propose a frame processing method that can perform well even with different video resolutions, using the Sign2Text method so that it can be applied to datasets without gloss. The overall overview of this paper follows Figure \ref{fig-label1}, this paper is summarized as follows: 

\begin{description}
  \item[$\bullet$]  We propose a new normalization method that is best suited for sign language translation in keypoint processing.
  \item[$\bullet$]  We propose a SASS method for extracting 'key frame' considering the characteristics of videos with different lengths.
  \item[$\bullet$]  Our method is a highly versatile methodology as it can be applied to datasets without glosses.
\end{description}
  
The rest of this paper is organized as follows: 
In Section 2, we survey the tasks of sign language translation and video processing methods.
In Section 3, we introduce our video processing method and SLT model. 
In Section 4, experiments demonstrate how effective our method is.
In Section 5, we analyze the effect of each method on performance through ablation experiments.
In Section 6, we share the qualitative results of applying our method.
Finally, Section 6 discusses the possibility of development of this study through conclusions and future work.

\begin{figure*}[h]
    \centering
    \includegraphics[width=1\linewidth]{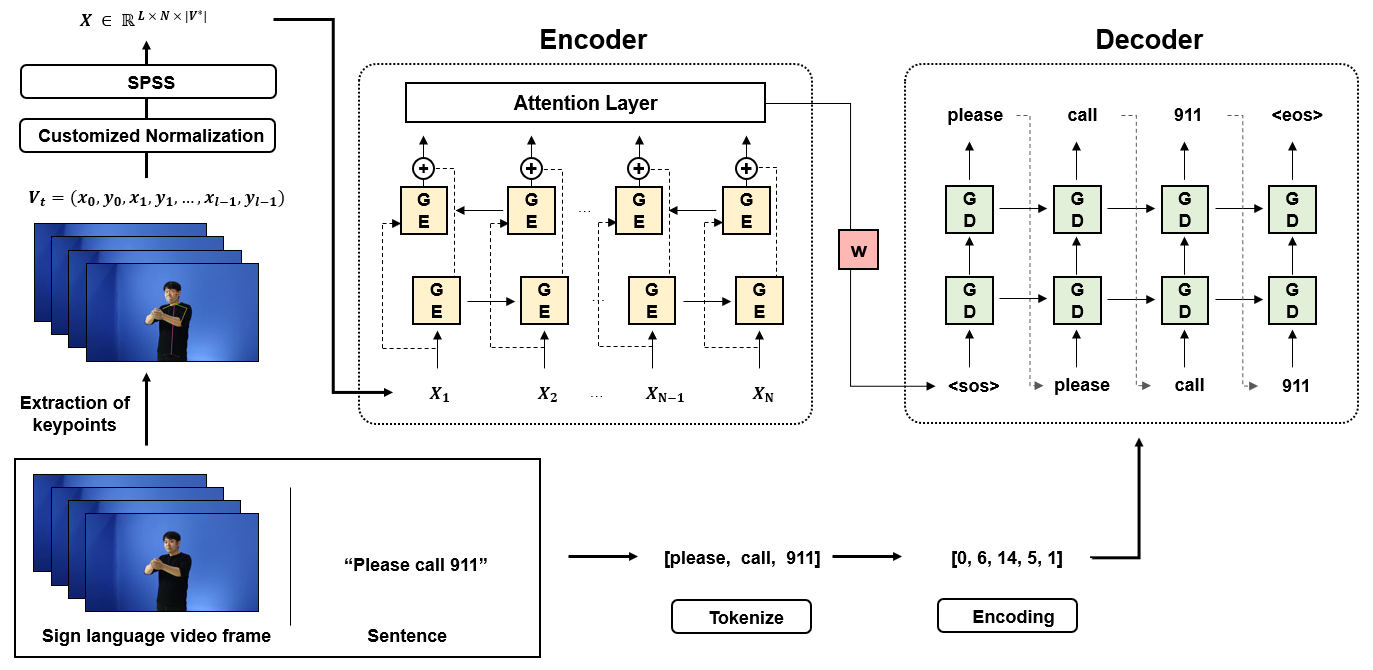}
    \caption{\small \textbf{A Full Architecture our SLT approach.} We extract keypoints from each frame and then go through the process of normalizing features. Then, after adjusting the number of frames, encoding is performed. And after tokenizing the text, it is put into the decoder through the embedding process. In this case, GE is bi-GRU-Encoder, GD is GRU-Decoder. And X is encoded vector that enters the model.}
    \label{fig-label2}
\end{figure*}

\section{Related Work}

\subsection{Sign Language Translation}
As we mentioned earlier, research on sign language translation has been steadily conducted.
However, there are several reasons for the slow development of the SLT system.
First, there is not enough data for sign language translation.
Sign language is used differently depending on each country and region, so the data that must be used when translating a language is different.
Therefore, it was necessary to collect sign language datasets by country directly, which was difficult because of the large cost.
In addition, previous studies only dealt with translation between sign language and text, and even though video information was not included, the average of total number of words was very small, about 3,000 \cite{fifteen, sixteen, seventeen}.
However, as algorithms developed, they laid the foundation for building sign language datasets.
In particular, with the development of algorithms for weakly announcement data \cite{eighteen,nineteen} and the development of algorithms for pose estimation \cite{twenty, twentyone}, researchers felt the need for sign language data and began to build datasets.

This development led to the creation of German sign language datasets RWTH-PHOENIX-Weather 2012 \cite{twentytwo}, RWTH-PHOENIX-Weather 2014 \cite{twentythree}, and RWTH-PHOENIX-Weather 2014-T \cite{thirteen} as well as the American sign language dataset ASLG-PC12 \cite{twentyfour} and Korean sign language dataset KETI \cite{three}.

With the construction of various datasets, the SLT task has developed through Deep Learning. In many CSLR or SLT studies, the frame is encoded in feature form through Convolution Neural Networks (CNNs) \cite{thirtysix}.
Camagoz \textit{et al}. \cite{thirteen} extracted features of frames using CNN and proposed an SLT method through the Encoding-Decoding method \cite{thirtyseven} using Recurrent Neural Networks (RNN).
Similarly, Zhou \textit{et al}. \cite{twentyfive} proposed a CSLR model that combines CNN and RNN. 

As Transformer \cite{twentysix} was in the spotlight in the NMT task, many SLT models using Transformer were developed.
Yin \textit{et al}. \cite{twentyseven} achieved the state of the arts by applying Transformer to the Zhou \textit{et al}. \cite{twentyfive} model called STMC, and Camagoz \textit{et al}.\cite{twentyeight} suggest a method of applying Transformer to \cite{thirteen} instead of RNN.
In addition, there is a study that proposed a BERT-based SLT method using a pretrained Transformer, unlike the existing aspect \cite{twentynine}.

There is also a skeleton-based SLT study using a model of the pose estimation task, not CNN. 
Ko \textit{et al}. \cite{three} achieved keypoint-based SLT on large-scale videos using KETI data sets. 
In addition, Jang \textit{et al}. \cite{thirty} perform translation using the Graph Convolution Network (GCN) \cite{fiftynine} to extract and weight important clips. Skeleton-based research continues not only in SLT but also in the task of CSLR \cite{thirtyone,thirtytwo,thirtythree}.

\subsection{Video Processing}

Video processing is mainly performed in action recognition tasks and video recognition tasks, which greatly influence performance \cite{forty,fortyone,fortytwo}.
Gowda \textit{et al}. \cite{thirtynine} has proposed a score-based frame selection method called SMART in the task action recognition, achieving state-of-the-arts for the UCF-101 dataset\cite{fortythree}, and Karpathy \textit{et al}. \cite{fortyfour} used a video processing method that treats all videos as fixed clips. 
Therefore, video processing is a very useful way to improve performance in a variety of tasks.

Video processing mainly utilizes a method of changing the number of frames, using a method of increasing the number of frames by converting the frames themselves \cite{fortysix,fifty}, or an augmented method of manipulating the order of frames \cite{fiftyone}. Recently, video processing is performed using Deep Learning, and there is also a study that proposed a frame augmentation method using Generative Advertising Networks (GAN) \cite{fortyfive} or using a genetic algorithm \cite{fortyseven}.

The frame processing method was used not only in the Action Recognition task but also in the SLT task.
Park \textit{et al}. \cite{fourteen} augmented the data in three ways: angle conversion of the camera, finger length conversion, and random keypoint removal.
In addition, Ko \textit{et al}. \cite{three} used a method of increasing the number of data by adding randomness to the new skip sampling method.

\begin{table}
\centering
\caption{\small \textbf{List of keypoint.} We divided it into four parts except for hands. The number is the index number of the corresponding part, and the reference point is the body part that is the reference for each part.}
\label{table1}
\resizebox{\columnwidth}{!}
{%
\begin{tabular}{c|c|c} 
\toprule
\textbf{Part}       & \textbf{Reference point} & \textbf{Included Number}  \\ 
\hline\hline
Face       & Nose (0)        & 0,1,2,3,4,11     \\
Upper Body & Neck (12)       & 5,6,12           \\
Left Arm   & Left Elbow (7)  & 7,9              \\
Right Arm  & Right Elbow (8) & 8,10             \\
\bottomrule
\end{tabular}
}
\label{table1}
\end{table}

\section{Proposed Approach}
In this section, we introduce the overall architecture of the proposed method. 
Our architecture is decomposed into four parts: Extract of keypoint, keypoint normalization, SASS and sign language translation model.
The overall flow of our proposed full architecture is described in Figure \ref{fig-label2}.

Camagoz \textit{et al}. \cite{thirteen,twentyeight} underwent the embedding process after extracting the feature of the sign language video frame through CNN. 
It does the aims to do the dimension reduction by embedding the feature point of the large amount extracted from the learning image.
However, all the features extracted can include backgrounds, which can affect learning by becoming noise.
Therefore, in this paper, instead of extracting features through CNN, we propose a method of extracting the keypoint of the signer and using it as a feature value.
Using keypoint, we can construct a strong model against the background by only looking at the movement of the signer.

We propose a video processing method for sign language video processing, which emphasizes the ‘key frame’ of the video so that the model can better learn important parts of the video.
First, we propose a “Customized Normalization” that further emphasizes important hand movements in sign language, considering the location of the keypoint.
In addition, we propose a stochastic augmentation method that gives the "key frame" of the video a stochastic weight so that the model can better learn important parts of the sign language video.
To apply not only frame augmentation but also dynamic video frame length $T$, "SASS" is proposed, which is a mixture of stochastic augmentation and skip sampling.
Finally, the sign language translation is achieved by applying the method of Bahdanau \textit{et al} \cite{thirtyseven} for the NMT task.
\begin{figure}
    \centering
    \includegraphics[width=1\columnwidth]{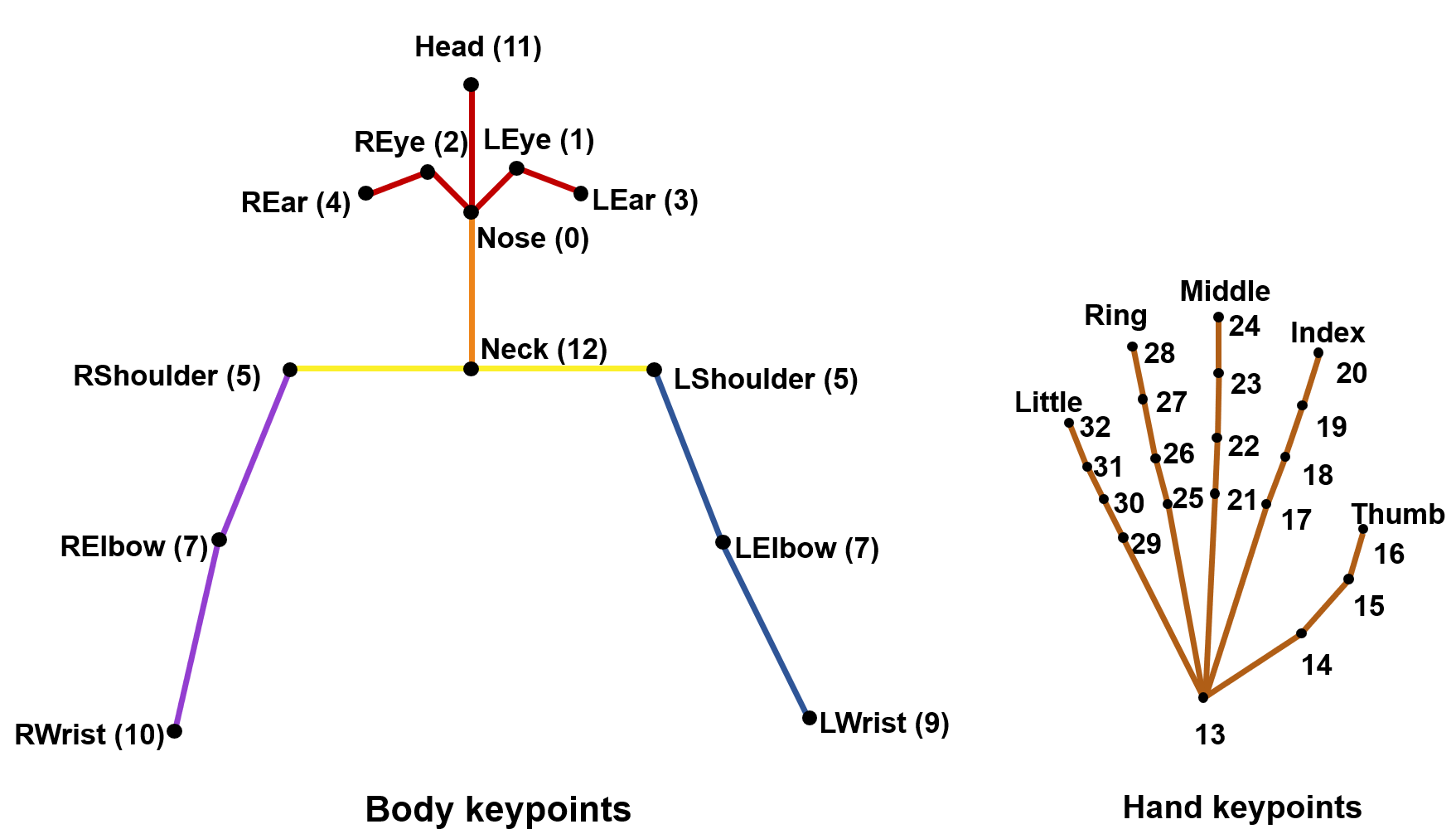}
    \caption{\small \textbf{Location and number for the keypoints.} We used 55 keypoints and excluded the keypoints of the lower body.
    Hand keypoints used both left and right hands. Then, we divided each normalization part by color.}
    \label{fig-label3}
\end{figure}
\subsection{Extraction Of Keypoint}
We extract keypoint of the signer to obtain the feature value of the video.
This method extracts skeletons points from the signer and minimizes the impact on the surrounding background, not the signer, in the video. 
We used a framework called Alphapose of Fang \textit{et al} \cite{twentyone}.
Alphapose can extract keypoints faster than the Cao, Zhe, \textit{et al} OpenPose \cite{twenty} model by first detecting signers in a top-down manner and then extracting keypoints from cropped images.
We used pretrained model on a Halpe dataset \cite{twentyone}. 
There are 136 keypoints in total, and we used 123 keypoints by removing 13 keypoints to exclude the lower body.
And 55 keypoints were used, excluding 68 detailed face keypoints, based on the better performances obtained when the keypoint of the face was removed in the study of Ko \textit{et al} \cite{three}.
Therefore, the keypoint vector is composed of $V = (v_{1},v_{2}, \ldots v_{55})$ and $v$ is the position coordinate $v\ _i\ =\left(v\ _i^x\ ,v\ _i^y\ \right)\ \ \ where\ \ \ i\ \in\ \left[1,55\right]$ of each keypoint. 
The number of the location of the keypoint follows in Figure \ref{fig-label3}.

\subsection{Customized Normalization}

In this paper, we propose customized normalization according to body parts considering keypoint positions.
Since the body length varies depending on the person, a normalization method considering this is needed.
Although the Robust Keypoint Normalization method was mentioned in Kim \textit{et al}. \cite{twelve}, it cannot be said to be a normalization method considering all positions because the reference point was set in some parts, not all parts.
Therefore, this paper proposes a normalization method that can further emphasize not only the location of each keypoint but also the movement of the hand.

We designated each reference point according to the part of the body the face, upper body, left arm, and right arm.
The reference point is defined as $r=(r_{x},r_{y})$.
The corresponding keypoint numbers according to each body part are shown in Table \ref{table1}, and the names corresponding to each number can be referred to in Figure \ref{fig-label3}.
In addition, a representative value for the entire body of the signer is defined as center point $c = (c_{x},c_{y})$ and follows (1).

\begin{equation}\label{eq}
  \begin{gathered}
    c_x\ =\ \frac{1}{55}\sum_{i=1}^{55}{\ v_i^x},\ c_y\ =\ \frac{1}{55}\sum_{i=1}^{55}v_i^y\  \\
  \end{gathered}
\end{equation}

In this case, $v^{x}_{i}\ \ , \ v^{y}_{i}$ are values of $V_{x} \ , \ V_{y}$ vectors separated by $x$ and $y$ coordinates concerning the vector $V$.
$V_x$ and $V_y$ are defined as follows:
\begin{equation}\label{eq}
  \begin{gathered}
    V_x\ =\ \left(v_1^x,v_2^x,\ \cdots,\ v_{55}^x\right) \\
    V_y\ =\ \left(v_1^y,v_2^y,\ \cdots,\ v_{55}^y\right)
  \end{gathered}
\end{equation}

We make the elements of vectors $V_{x}$ and $V_{y}$ follows the (3). 
\begin{equation}\label{eq}
    \begin{gathered}
        S_{point}\ =\ \{\ Nose,\ Neck,\ LElbow,\ RElbow\ \} \\
        d_{point}\ =\ \sqrt{\left(c_x\ -\ r_x\right)^2\ +\ \left(c_y-r_y\right)^2}\\
        where\ \left(r_x,r_y\right)\ \in S_{point} \\
        \left(\ \widetilde{\ x}_{body},\ \widetilde{\ y}_{body}\ \right)\ =\ \left(\frac{v^x-c_x}{d_{point}},\ \frac{v^y-c_y}{d_{point}}\right)
    \end{gathered}
\end{equation}

Equation (3) performs normalization based on the distance $d_{point}$ after obtaining the euclidean distance ${d_{point}}$ between vector $c$ and the reference point according to each part.
Therefore, the keypoint for the body excluding both hands follows $\left(\widetilde{x}_{body},\ \widetilde{y}_{body}\right)$.

Next, we perform MinMax scaling to make the keypoints of both hands equal in scale and compare the movements of both hands equally.
MinMax Scaling is to adjust the range to [0,1] by dividing the difference between the maximum and minimum values of coordinates by subtracting the corresponding coordinates and the minimum values.
Here, to prevent a situation in which the minimum value becomes 0 and the vanishing gradient problem occurs, we subtracted -0.5 and adjusted it to the range of [-0.5,0.5]. This follows (4).
Therefore, we define the keypoints of both hands that have undergone this process as $\widetilde{x}_{hand}, \widetilde{y}_{hand}$.
This follows (4):

\begin{equation}\label{eq}
    \begin{gathered}
    \left(\widetilde{x}_{hand}\ ,\ \widetilde{y}_{hand}\right) \\
    =\ (\frac{v^x\ -\ v_{min}^x}{v_{max}^x\ -\ v_{min}^x}\ -\ 0.5\ , \frac{v^y\ -\ v_{min}^y}{v_{max}^y\ -\ v_{min}^y}\ -\ 0.5\ )
    \end{gathered}
\end{equation}
where $v^{x}_{max}$  and $v^{x}_{min}$ are the maximum and minimum values of the hand keypoint, respectively.
Thus, each $x$ and $y$ point to which our normalization method is applied is defined as vectors $V_x^*\ =\ \left[\widetilde{x}_{body};\widetilde{x}_{hand}\right]$ and $V_y^\ast\ =\ \left[\widetilde{y}_{body};\widetilde{y}_{hand}\right]$ and the final keypoint vector becomes $V^*\ =\ \left[V_x^*;V_y^*\right]\in \mathbb{N}^{{55}\times2}$.

\begin{figure}[t]
    \centering
    \includegraphics[width=1\linewidth]{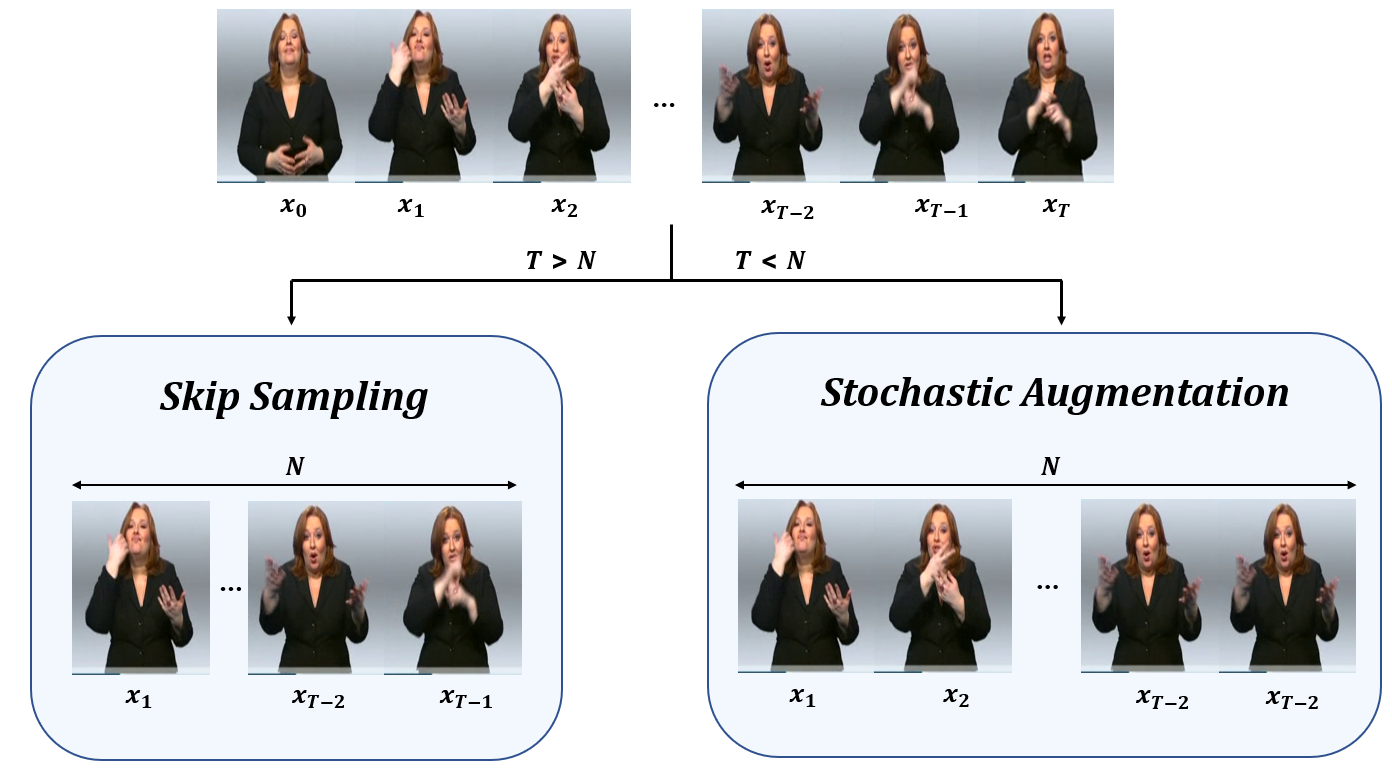}
    \caption{\small \textbf{SASS} This is our frame selection method. If the length $T$ of the video frame is less than a fixed scalar $N$, a stochastic augmentation method is used, and if $T$ is greater than $N$, skip sampling is used to match $T$ with the number of $N$.}
    \label{fig-label4}
\end{figure}

\subsection{SASS}

To train through deep learning, the input size of the model must be fixed.
In previous studies, for this purpose, the dimension was reduced and sized through embedding \cite{thirteen,twentyeight}.
However, this results in a loss of information in the video. Therefore, we use a method that reflects the information about the sign language video as it is without using embedding.
To this end, this study uses a method of fixing the length of the input value by augmenting or sampling the frame.

In this paper, we propose a general-purpose method that fixes the length of the input value but applies to various datasets.
In addition, we emphasize the hand-moving frame, which is the key frame of the sign language video, and propose a method of not losing it.
Since the frames existing at the beginning and end of the video are those with less hand movement of most signers, we propose an augmentation methodology that emphasizes the middle frame of the image except for this.
And, when the video length is very long, a sampling method that does not lose the key frame is applied.
In this paper, combining these two methods, we propose an augmentation and sampling method that takes into account the difference in frames for each video.
We call this method "Stochastic Augmentation and Skip Sampling (SASS)".

We define the $i$-th training video as $\textsc{x}_{i}\ =\ \{\ x_t\ \}_{t=0}^T$ and $L$ as the number of training videos.
where, $i \in [1,L]$.
We select N frames from between the frames of the training video $\textsc{x}_i$, which follows (5).

\begin{equation}
    N\ =\ \frac{1}{L}\ \sum_{i=0}^{L}\textsc{x}_i
\end{equation}
 
If the number of video frames $\textsc{x}_{i}$ is greater than $N$, the sampling is used, and if the number of video frame is less than $N$, the number of frames in each video is adjusted to $N$ using the augmentation. 
So, input value $X$ is $X \in \mathbb{R}^{L \times N \times |V^*|}$ 
This method can prevent the loss of video information, which is a disadvantage of frame sampling, and the disadvantage of frame augmentation, which can prevent a lot of memory consumption. 
The overall flow of our SASS method follows Figure \ref{fig-label4}.

\subsubsection{Stochastic Augmentation}

Our proposed method is not to follow a uniform probability when randomly selecting a frame, but to reconstruct a key-frame into a probability distribution that can be preferentially extracted with a high probability of being selected.
The probability set in which the frame is selected through the change in the probability $p$ based on the binary distribution is constituted.
We define a set of frame selection probabilities with the probability of $p$ as $F_{p}$.
Therefore, $F_{p}$ is defined as $F_{p} = \{f(0;T,p), f(1;T,p), \ldots , f(T-1;T,p)\}$ where function $f$ follows as (6):
\begin{equation}
    f(k;T,p) = {T-1 \choose k} p^{k} (1-p)^{T-k-1} \ \ where \ \ k \in [0,T-1]
\end{equation}
Key frames may be preferentially selected for each frame through the probability of having such a binomial distribution form.

We do not fix p to a single value but construct a set $Pr$ to construct several probability values.
The set $Pr_{n}$ follows (7).
\begin{equation}\label{eq}
    \begin{gathered}
        Pr_{0} = \{ \frac{1}{2} \} \\
        Pr_{n} = Pr_{n-1} \cup \{ \frac{1}{n+2}, \frac{n+1}{n+2} \} \\
        p \sim Pr_{n}
    \end{gathered}
\end{equation}

That is, the probability $p$ follows the probability set $Pr_n$, and accordingly, $F_p$ is reconstructed.
The reconstructed set of frame selection probabilities is called $F^{*}_{p}$, which follows (8).
\begin{equation}\label{eq}
    F^{*}_{p} = \frac{1}{l_{p}} \sum_{p \in Pr_{n}}F_{p}
\end{equation}

where,$l_p$ is length of $Pr_n$
However, constructing $F^{*}_{p}$ in this way does not augment the middle portion of the video, which is the original purpose of the probability.
Therefore, we rearrange $F^{*}_{p}$ based on the median to increase the probability of selection for the middle portion of the video.
\begin{figure}[t]
    \centering
    \includegraphics[width=1\linewidth]{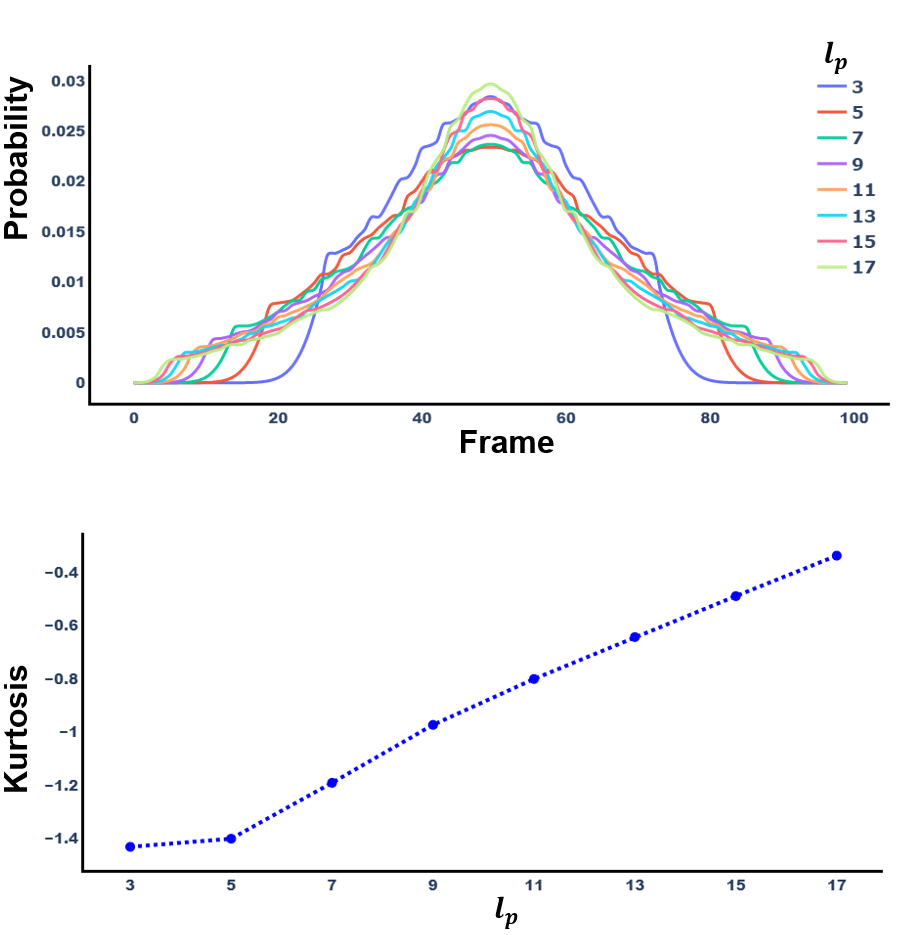}
    \caption{\small \textbf{Change of probability distribution and kurtosis following $n$.} 
    (Top) Probability distribution of probability set $F^{*}_n$ according to the change in $l_p$.
    (Bottom) Kurtosis of $F^{*}_n$ according to the change in $l_p$.}
    \label{fig-label5}
\end{figure}
That is, if the frame order $k$ is smaller than $\lfloor \frac{T}{2} \rfloor$, it is sorted in ascending order, and if it is larger, it is sorted in descending order.
Accordingly, a final set of frame selection probabilities is produced, and based on this, priority probabilities for frame augmentation order are obtained.
Figure \ref{fig-label5} shows the probability distribution and kurtosis of $F^{*}_{p}$ according to $l_p$. As $l_p$ increases, kurtosis increases.

\begin{table*}[t]
\centering
\caption{\textbf{Comparison Of Normalization Method}}
\resizebox{\textwidth}{!}{%
\begin{tabular}{c|cc|cc|cc|cc|cc|cc} 
\hline
\multirow{3}{*}{Method} & \multicolumn{6}{c|}{KETI}                                                                          & \multicolumn{6}{c}{RWTH-PHOENIX-Weather
  2014 T}                                                   \\ 
\cline{2-13}
                        & \multicolumn{2}{c|}{BLEU-4}    & \multicolumn{2}{c|}{ROUGE-L}    & \multicolumn{2}{c|}{METEOR}     & \multicolumn{2}{c|}{BLEU-4}     & \multicolumn{2}{c|}{ROUGE-L}    & \multicolumn{2}{c}{METEOR}      \\ 

                        & dev           & test           & dev            & test           & dev            & test           & dev            & test           & dev            & test           & dev           & test            \\ 
\hline
Standard                & 78.99         & 76.44          & 78.84          & 76.28          & 78.84          & 76.28          & 12.08          & 7.36           & 24.74          & 23.57          & 26.93         & 25.87           \\
Robust~                 & 79.17         & 76.29          & 79.07          & 76.23          & 79.08          & 76.18          & 9.19           & 8.04           & 24.73          & 23.57          & 26.92         & 25.86           \\
MinMax                  & 79.75         & 79.01          & 79.75          & 78.9           & 79.75          & 78.9           & 9.88           & 8.05           & 26.04          & 24.86          & 27.47         & 26.11           \\
Kim et al\cite{twelve} method    & 73.95         & 72.91          & 73.8           & 72.77          & 73.8           & 72.77          & 11.67          & 10.45          & 27.2           & \textbf{26.22} & 27.92         & 27.2            \\
All reference           & 77.98         & 76.35          & 78.09          & 76.22          & 78.06          & 76.22          & 11.16          & 9.3            & 27.08          & 25.68          & 28.22         & \textbf{27.33}  \\
Center Reference        & 75.65         & 75.02          & 75.82          & 74.81          & 75.79          & 74.81          & 12.1           & 8.28           & \textbf{27.66} & 25.68          & \textbf{28.8} & 26.79           \\
\hline
Ours              & \textbf{85.3} & \textbf{84.39} & \textbf{85.48} & \textbf{84.85} & \textbf{85.58} & \textbf{82.31} & \textbf{12.81} & \textbf{13.31} & 24.64          & 24.72          & 25.86         & 25.85           \\
\hline
\end{tabular}%
}
\label{table2}
\end{table*}

\subsubsection{Skip Sampling}

We chose a sampling method to avoid missing key frames when selecting a frame. 
The sampling method of Ko \textit{et al} proposed a random sampling method that does not lose the key frame. 
Therefore, this paper also conducts sampling according to this method.
When a fixed size is $N$ and the number of frames in the current video is $T$, the average difference $z$ of frames follows the following (9).
\begin{equation}
    c = \lfloor \frac{T}{N-1} \rfloor
\end{equation}
$S_N$ is the framing sequence of the video in the form of an arithmetic sequence and is defined as shown in (10) using $z$.
\begin{equation}
    \begin{gathered}
        S_N = s + (N-1)c \ \ where\ \ s = \frac{T - c(N-1)}{2} \\
        \{ S_N \}_{N \in \mathbb{N}^N}
    \end{gathered}
\end{equation}

Random sequence $R$ with a sequence range of $[1,c]$ is added to the baseline sequence $S_{N}$ defined in this way.
Note that the value of the last index is clipped to the value in the range of $[1, T]$.
If skip sampling is defined in this way, the order of frames can be considered without losing the key frame.

\subsection{Sign Language Translation}
We used a GRU-based model with a sequence to sequence (Seq2Seq) structure with an encoder and decoder in the translation step and improved translation performance by adding attention.
This is a structure widely used in NMT and is suitable for translation tasks with variable-length inputs and outputs.
We used Bahdanau attention \cite{thirtyseven}.
The encoder-decoder structure goes through the following process when source sentence $\textsc{x} = (x_{1},x_{2}, \ldots , x_{T_{x}} )$ and target sentence $\textsc{y} = (y_{1},y_{2}, \ldots , y_{T_{y}} )$. 
Here, our model encodes a keypoint vector, not a sentence.
First, the encoder calculates the hidden state and generates a context vector through it, and equation follows:
\begin{equation}
    \begin{gathered}
    h_{t} = f(x_{t}, h_{t-1}) \\
    c = q(h_{1}, \ldots , h_{T_{x}})
    \end{gathered}
\end{equation}
$h_t$ is the hidden state calculated in time step $t$ and is the context vector generated through the encoder and nonlinear function.
Next, in decoder, the joint probability can be calculated in the process of predicting words. This follows (12).
\begin{equation}
    p(\textsc{y}) = \prod^{T_{y}}_{i=1}p(y_{i}| \{ y_{1},y_{2}, \ldots , y_{i-1} \} , c )
\end{equation}
The conditional probability at each time used in the (12) can be calculated through the following equation.
\begin{equation}
    p(y_{i}| y_{1},y_{2}, \ldots , y_{i-1} , \textsc{x} ) = softmax(g(s_{i}))
\end{equation}
And in (13), $s_{i}$ is the hidden state calculated from the time $i$ of the decoder and follows the following equation.
\begin{equation}
    s_{i} = f(y_{i-1},s_{i-1},c)
\end{equation}
The context vector $c_{i}$ is calculated using a weight for $h_{i}$ and follows (15).
\begin{equation}
    c_{i} = \sum^{T_{j}}_{j=1} = a_{ij}h_{j}
\end{equation}

where,
\begin{equation}
    a_{ij} = \frac{ exp(score(s_{i-1},h_{k})) }{\sum^{T_x}_{k=1} exp(score(s_{i-1},h_{k})}    
\end{equation}

Here, score is an alignment function that calculates the match score between the hidden state of the encoder and the decoder. 
\begin{table*}[t]
\centering
\caption{\small (TOP) \textbf{Comparison of SLT Performance by Sampling Method} \\(Bottom) \textbf{Comparison of SLT Performance by Augmentation Method}}    
\resizebox{\textwidth}{!}{%
\begin{tabular}{c|cc|cc|cc|cc|cc|cc} 
\hline
\multirow{3}{*}{Method}   & \multicolumn{6}{c|}{KETI}                                                                           & \multicolumn{6}{c}{RWTH-PHOENIX-Weather
  2014 T}                                                                                      \\ 
\cline{2-13}
                          & \multicolumn{2}{c|}{BLEU-4}     & \multicolumn{2}{c|}{ROUGE-L}    & \multicolumn{2}{c|}{METEOR}     & \multicolumn{2}{c|}{BLEU-4}     & \multicolumn{2}{c|}{ROUGE-L}    & \multicolumn{2}{c}{METEOR}                                         \\ 

                          & dev            & test           & dev            & test           & dev            & test           & dev            & test           & dev            & test           & dev            & test                                              \\ 
\hline
Random
  sampling         & 76.09          & 75.11          & 77.17          & 16.61          & 77.08          & 76.45          & 7.77           & 5.94           & 21.74          & 21.12          & 23.36          & 22.63                                             \\
Skip Sampling             & \textbf{82.4}  & \textbf{81.9}  & \textbf{83.35} & \textbf{82.74} & \textbf{83.32} & \textbf{82.96} & \textbf{9.52}  & \textbf{7.55}  & \textbf{25.07} & \textbf{22.35} & \textbf{26.36} & \textbf{23.77}                                    \\
Stochastic
  Sampling     & 78.9           & 78.2           & 79.92          & 79.07          & 79.78          & 79.08          & 8.19           & 6.44           & 21.45          & 21.29          & 22.82          & \begin{tabular}[c]{@{}c@{}}22.86 \\\end{tabular}  \\ 
\hline \hline
Random
  Augmentation     & \textbf{83.69} & \textbf{83.68} & \textbf{84.19} & \textbf{84.15} & \textbf{84.19} & \textbf{84.25} & 13.71          & 11.3           & \textbf{28.11} & 27.36          & \textbf{27.59} & \textbf{26.89}                                    \\
Stochastic
  Augmentation & 82.25          & 83.33          & 83.19          & 80.06          & 80.74          & 80.75          & \textbf{13.74} & \textbf{12.65} & 28.08          & \textbf{27.59} & 26.36          & \begin{tabular}[c]{@{}c@{}}25.73 \\\end{tabular} 
\end{tabular}%
}
\label{table3}
\end{table*}

\begin{table*}
\centering
\caption{\textbf{Comparison of combinations of sampling and augmentation}}
\resizebox{\textwidth}{!}{%
\begin{tabular}{c|cc|cc|cc|cc|cc|cc} 
\hline
\multirow{3}{*}{}         & \multicolumn{6}{c|}{KETI}                                                                          & \multicolumn{6}{c}{RWTH-PHOENIX-Weather
  2014 T}                                                                                              \\ 
\cline{2-13}
                          & \multicolumn{2}{c|}{BLEU-4}    & \multicolumn{2}{c|}{ROUGE-L}    & \multicolumn{2}{c|}{METEOR}     & \multicolumn{2}{c|}{BLEU-4}     & \multicolumn{2}{c|}{ROUGE-L}    & \multicolumn{2}{c}{METEOR}                                                 \\ 

                          & dev           & test           & dev            & test           & dev            & test           & dev            & test           & dev            & test           & dev            & test                                                      \\ 
\hline
Skip + Random             & 84.88         & 84.24          & 85.25          & 84.75          & 85.29          & 84.89          & 11.64          & 11.5           & 27.63          & \textbf{27.63} & 29.69          & 28.68                                                     \\
Stochastic +
  Random     & 83.85         & 83.92          & 84.44          & 84.65          & 84.4           & 84.57          & 10.36          & 10.16          & 27.48          & 26.77          & 29.47          & 28.38                                                     \\
Stochastic +
  Stochastic & 84.94         & 83.5           & 85.37          & 83.93          & 85.45          & 83.95          & 9.71           & 10.83          & \textbf{27.88} & 27.04          & \textbf{30.01} & \begin{tabular}[c]{@{}c@{}}\textbf{28.7 }\\\end{tabular}  \\ 
\hline
SASS (ours)               & \textbf{85.3} & \textbf{84.39} & \textbf{85.48} & \textbf{84.85} & \textbf{85.58} & \textbf{85.07} & \textbf{12.81} & \textbf{13.31} & 24.64          & 24.72          & 25.86          & \begin{tabular}[c]{@{}c@{}}25.85 \\\end{tabular}         
\end{tabular}%
}
\label{table4}
\end{table*}
\section{Experiments}

To prove the effectiveness of our proposed method, we conduct several experiments on two diverse datasets: KETI\cite{three}, RWTH-PHOENIX-Weather 2014 T\cite{thirteen}.
KETI dataset is a Korean sign language video consisting of 105 sentences and 419 words and has a full high definition (HD) video.
There were 10 signers, and they filmed the video from two different angles.
However, the KETI dataset is not provided with a test set, so we randomly split the dataset at a ratio of 8:1:1.
The split of videos for train, dev and test is 6043, 800 and 801, respectively.
We divided these videos into frames at 30 fps.
KETI dataset did not provide gloss, so it immediately translated to the spoken language.

The RWTH-PHOENIX-Weather-2014-T dataset is a data set extended from the existing RWTH-PHOENIX-Weather 2014\cite{twentythree} with the public german sign language.
The split of videos for train, dev and test is 7096, 519 and 642, respectively. 
It has no overlap with the RWTH-PHOENIX-Weather 2014.

The average number of frames of training videos in these two datasets are 153 and 116 frames, respectively.
Therefore, the input X entering our model follows $X_{kor} \in \mathbb{R}^{6403 \times 153 \times 110}$ and $X_{ger} \in \mathbb{R}^{6403 \times 116 \times 110}$, respectively.

Our experiment was conducted in NVIDIA RTX A6000 and AMD EPYC 7302 16-Core environment for CPU. Our model was constructed using Pytorch\cite{fiftytwo}, Adam Optimizer\cite{fiftythree}, and Cross-Entropy loss.
The learning rate was set to 0.001 and epochs 100, and the dropout ratio was set to 0.5, preventing overfitting.
Finally, the dimension of hidden states was 512.

We tokenize differently according to the characteristics of Korean and German.
In Korean, the KoNLPy library's Mecab part-of-speech (POS) tagger was used, and in German, the tokenization was performed through the nltk library \cite{fiftyfive}. Finally, our evaluation metric evaluates our model in three ways: BLEU-4\cite{fiftysix}, ROUGE-L\cite{fiftyseven}, and METEOR\cite{fiftyeight} according to the metric of NMT.

We experiment on a total of four things. 
First, a comparative experiment according to the normalization method is conducted.
Second, an experiment on video sampling and augmentation is conducted.
Third, we experiment with changes in the length of set $Pr_{n}$(i.e. $l_p$), the representative value $N$, and the sequence of the input value.
Finally, we demonstrate the excellence of our model compared to previous studies.

\subsection{Effect Of Normalization}

This experiment proves that our method is powerful compared to various normalization methods.
The results of this are shown in Table \ref{table2}.
We fixed and experimented with all elements except normalization method.
We compared our method with the method called "Standard Normalization($X = (x- \mu)/\sigma$)" using the standard deviation($\sigma$) and average value($\mu$) of points, the method called "Min-Max Normalization($X = (x - x_{max})/(x_{max} - x_{min}) - 0.5$)" using the minimum($x_min$) and maximum values($x_{max}$) of points and the method called "Robust Normalization($X = (x - x_{2/4})/(x_{3/4} - x_{1/4})$)" using the median($x_{2/4}$) and interquartile range (IQR).
where, IQR is $x_{3/4} - x_{1/4}$.
In this case, standard normalization is the method proposed in Ko \textit{et al}.

In addition, we experimented by adding two normalization methods that did not exist before.
First, "all reference" is a normalization method in which (3) is applied to the hand.
Second, "center reference", is a method using only each keypoint vector $v$ and center vector $c$ without a separate reference point.
This divides the difference between v and c by the distance.
Finally, the reference point was compared with the Kim \textit{et al}\cite{twelve} method of normalization by fixing it with the right shoulder.

Our method proved to be the most powerful method through the experiment.
It can be seen that the performance has improved a lot compared to the methods of Kim \textit{et al} and Ko \textit{et al}, which are previous studies.
In addition, it has proven that it is a new powerful method that can improve performance by performing better not only in the dataset (KETI) with a relatively large scale frame but also in PHOENIX with a relatively small image.

\subsection{Effect Of SASS}
In this section, a difference in performance according to a frame selection method is experimented.
Keypoints were normalized with our experimental normalization method, and everything else was fixed except the frame selection method.
We demonstrate that our proposed method is excellent with a total of three comparative experiments.
We conduct comparison experiments using only sampling, comparison experiments using only augmented, and comparison experiments using both sampling and augmentation methods.

First, we compare the experimental results when only sampling is performed.
As the sampling method, the stochastic frame sampling and skip sampling methods proposed in this paper are compared.
It is also compared with a random sampling technique that randomly selects a simple equal distribution without such additional processing.

In this case, the number of sampling was set to $N = 50$ for the KETI data set and compared to the Ko \textit{et al} method, and in the PHOENIX data set, the minimum frame length $N = 16$ of the video was set.
In the video sampling comparison experiment, the skip sampling technique showed the best results for the two datasets. In addition, probability sampling also differs significantly compared to random sampling following an even distribution.

Second, we experimented with the performance change when only frames were augmented.
We set the maximum frame value of the training video in the dataset to $N$, and set the $N$ of KETI and PHOENIX to 426 and 475, respectively.
We compared and experimented with a randomly augmented method following an even distribution and a stochastic augmented method following our method.
This is shown in Table \ref{table3}.
\begin{table}
\centering
\caption{\small \textbf{BLEU-4 score by $l_p$}}
\begin{tabular}{c|cc|cc}
\multirow{2}{*}{$l_p$} & \multicolumn{2}{c|}{KETI} & \multicolumn{2}{c}{PHOENIX}                               \\ 
                          & dev   & test              & dev   & test                                              \\ 
\hline
3                         & 84.38 & 83.1              & 11.25 & 11.17                                             \\
5                         & 84.49 & 83.27             & 12.23 & 10.48                                             \\
7                         & 84.4  & 83.97             & 12.42 & 11.74                                             \\
9                         & 83.41 & 83.85             & 14.84 & 12.28                                             \\
11                        & 84.35 & 83.45             & 13.49 & 12.03                                             \\
13                        & 84.78 & 83.31             & 12.31 & 12.01                                             \\
15                        & 84.76 & 81.68             & 12.64 & 11.68                                             \\
17                        & \textbf{85.3}  & \textbf{84.39}             & \textbf{12.81} & \textbf{13.31}                                             \\ 
Mean                      & 84.48 & 83.38             & 12.75 & \begin{tabular}[c]{@{}c@{}}11.84 \\\end{tabular} 
\end{tabular}
\label{table5}
\end{table}
At KETI, the random augmentation was good in every way.
However, in PHEONIX, the main evaluation index, BLEU-4, was the best in stochastic augmentation.
The overall performance was improved when the frame augmentation was performed compared to when only sampling was performed.

Finally, we experimented with a combination of Sampling and Augmentation.
We experiment with a total of 4 cases.
In the previous experiment, since the random sampling method among the sampling methods has poor results, the experiment was conducted by combining the number of cases except random sampling.
This is shown in Table \ref{table4}.
The SASS method, which combines our methods of skip sampling and stochastic augmentation, performed well in both KETI, and BLEU-4 was overwhelmingly the highest in PHOENIX.

\subsection{Additional Experiments}
In this section, we experiment on several factors affecting the model.
We compare the performance according to the change in the probability set $Pr_n$ and compare according to $N$.

And experiments are conducted on how to invert the order of input values, a method of enhancing the translation performance proposed in the study of Sutskever \textit{et al} \cite{thirtyfive}.
For the first time, we experiment with the performance change according to $l_p$, the number of probability sets $Pr_n$.

We experiment only with $l_p \in [3,5,7,9,11,13,15,17]$ and evaluate only with BLEU-4.
This is shown in Table \ref{table5}.
We achieve the best performance on both datasets when $l_p$ is 17.
Even when compared to the average BLEU value, in the PHOENIX dataset, all of our methods are superior to other methodologies.
\begin{table}
\centering
\caption{\small \textbf{BLEU-4 score by $N$}}
\begin{tabular}{c|cc|cc}
\multirow{2}{*}{$N$} & \multicolumn{2}{c|}{KETI} & \multicolumn{2}{c|}{PHOENIX}  \\
                  & dev   & test              & dev   & test                  \\ 
\hline
Mean              & \textbf{85.3}  & \textbf{84.39}             & \textbf{12.81} & \textbf{13.31}                 \\
Median            & 85.22 & 82.78             & 12.65 & 9.27                 
\end{tabular}
\label{table6}
\end{table}

\begin{table}
\centering
\caption{\small \textbf{BLEU-4 score in order of input values}}
\begin{tabular}{c|cc|cc}
\multirow{2}{*}{Order} & \multicolumn{2}{c|}{KETI}      & \multicolumn{2}{c}{PHOENIX}                                                 \\ 
                       & dev           & test           & dev            & test                                                       \\ 
\hline
Reverse                & 85.1          & 84.3           & 12.38          & 9.13                                                       \\
Original               & \textbf{85.3} & \textbf{84.39} & \textbf{12.81} & \begin{tabular}[c]{@{}c@{}}\textbf{13.31 }\\\end{tabular} 
\end{tabular}
\label{table7}
\end{table}
Next, an experiment was conducted to compare according to the set criteria of $N$.
We set and compare $N$ according to the mean and median frame length of the training video.
The median number of frames of training videos in these two datasets are 148 and 112 frames, respectively.
The comparison accordingly is shown in Table \ref{table6}.

Finally, we experiment with the effect of encoding by inverting the order of input $X$.
This method takes into account that in a linguistic system, there is a higher association between the words located in the front, so that the match with the target sentence can be better if the input order of the source sentence is reversed.
Therefore, we reverse the order of frames and encode each video by changing the order of frames to $(x_{T-1}, x_{T-2}, ..., x_0)$.
This is shown in \ref{table7}.
This result shows that setting $N$ as the average has the best effect, and also that entering the original order of the frame has the best effect.
It is analyzed that the change in the order of the input sentences is meaningless because our input value is the keypoint value of the frame, not the sentence.

\subsection{Comparison Of Previous Works}
In this section, we prove the excellence of our model by comparing existing studies with our model.
KETI dataset was implemented directly and conducted a comparative experiment because we randomly divided it into the training, validation, and test sets.
In Ko \textit{et al}'s study, additional data was used, and we reproduced and compared the proposed model without adding data.
This is shown in Table \ref{table8}.

In addition, since we are a robust method of experimenting with a gloss-less dataset, PHOENIX dataset only compares video to text translation, not gloss-based methods.
This is shown in Table \ref{table9}.

Our model also performs well compared to previous studies in the dataset.
In particular, it can be seen that KETI-dataset exhibits the best performance among keypoint-based SLT models.
In addition, the BLEU score was the highest in the PHOENIX dataset. 
BLEU-4 increased significantly (about 4\%p) compared to Sign2Text (Bahdanadu Attention), indicating that the ideal performance was improved through the Our video processing method even though the same Bahadanadu Attention was used.

\begin{table}
\centering
\caption{\small \textbf{Comparison with other models in KETI dataset}}
\resizebox{\columnwidth}{!}{%
\begin{tabular}{c|cc|cc|cc}
              & \multicolumn{2}{c|}{BLEU-4} & \multicolumn{2}{c|}{ROUGE-L} & \multicolumn{2}{c}{METEOR}  \\
              & dev   & test                & dev   & test                 & dev   & test                \\ 
\hline
Ko et al\cite{three}   & 77.96 & 76.44               & 77.99 & 76.56                & 77.99 & 76.49               \\
Kim et al\cite{twelve} & -     & 78.00               & -     & 81.00                & -     & 83.00               \\
Ours  & \textbf{85.3}  & \textbf{84.39}               & \textbf{85.48} & \textbf{84.85}                & \textbf{85.58} & \textbf{85.07}              
\end{tabular}%
}
\label{table8}
\end{table}

\begin{table}
\centering
\caption{\small \textbf{Comparison with other models in PHOENIX dataset}}
\resizebox{\columnwidth}{!}{%
\begin{tabular}{c|cc|cc|cc}
                                                                                              & \multicolumn{2}{c|}{BLEU-4}     & \multicolumn{2}{c|}{ROUGE-L}  & \multicolumn{2}{c}{METEOR}       \\
                                                                                              & dev            & test           & dev           & test          & dev            & test            \\ 
\hline
\begin{tabular}[c]{@{}c@{}}Sign2text\\(Luong)\cite{thirteen}\end{tabular}    & 10.00          & 9.00           & 31.8          & \textbf{31.8} & -              & -               \\
\begin{tabular}[c]{@{}c@{}}Sign2Text\\(Bahdanau)\cite{thirteen}\end{tabular} & 9.94           & 9.58           & \textbf{32.6} & 30.7          & -              & -               \\
Ours                                                                                          & \textbf{12.81} & \textbf{13.31} & 24.64         & 24.72         & \textbf{25.86} & \textbf{25.85} 
\end{tabular}%
}
\label{table9}
\end{table}

\begin{table*}
\centering
\caption{\small \textbf{Ablation Study of Video Processing Methods.} "Normalization" means our customized normalization method. "Sampling" is our skip sampling method and "Augmentation" is our stochastic augmentation method.}
\resizebox{\textwidth}{!}{%
\begin{tabular}{ccc|ccc|ccc}
\multicolumn{3}{c|}{}                   & \multicolumn{3}{c|}{KETI}                        & \multicolumn{3}{c}{RWTH-PHOENIX-Weather-2014 T}                                     \\ 
\hline
Normalization & Sampling & Augmentation & BLEU-4         & ROUGE-L        & METEOR         & BLEU-4         & ROUGE-L        & METEOR                                            \\ 
\hline
              & \checkmark        &              & 76.44          & 76.56          & 76.49          & 7.48           & 21.91          & 23.75                                             \\
              &          & \checkmark            & 77.00          & 77.62          & 77.81          & 11.4           & 24.62          & 24.93                                             \\
              & \checkmark        & \checkmark            & 77.78          & 78.42          & 78.61          & 7.36           & 24.5           & \textbf{26.21}                                    \\
\checkmark             & \checkmark        &              & 81.9           & 82.74          & 82.96          & 7.55           & 22.35          & 23.77                                             \\
\checkmark             &          & \checkmark            & 80.06          & 80.74          & 80.75          & 12.65          & \textbf{27.59} & 25.73                                             \\
\checkmark             & \checkmark        & \checkmark            & \textbf{84.39} & \textbf{84.85} & \textbf{85.07} & \textbf{13.31} & 24.72          & \begin{tabular}[c]{@{}c@{}}25.85 \\\end{tabular} 
\end{tabular}%
}
\label{table10}
\end{table*}
\section{Ablation Study}
We conduct ablation studies to analyze the effectiveness of the our proposed method. 
We experimented with the change in performance depending on the presence or absence of our normalization and frame selection techniques.
We study the following seven components of our method. This is shown in Table \ref{table10}.
\begin{table}[t!]
\centering
\caption{\small \textbf{Translation from our networks} (GT : Ground Truth)}
\resizebox{\columnwidth}{!}{%
\begin{tabular}{c|l} 
\hline
\multicolumn{2}{c}{RWTH-PHOENIX-Weather-2014 T}                                                                                                                                         \\ 
\hline
\multirow{2}{*}{Reference} & \begin{tabular}[c]{@{}l@{}}und nun die wettervorherage \\ür morgen amtag den zweiten April .\end{tabular}                                                    \\
                           & \begin{tabular}[c]{@{}l@{}}(and now the weather forecast \\for tomorrow morning the second April .)\end{tabular}                                             \\ 
\hline
\multirow{2}{*}{Ours}      & \begin{tabular}[c]{@{}l@{}}und nun die wettervorhersage\\für morgen samstag den achtundzwanzigsten juli .\end{tabular}                                       \\
                           & \begin{tabular}[c]{@{}l@{}}(and now the weather forecast for\\tomorrow Saturday the twenty-eighth of July.)\end{tabular}                                   \\ 
\hline
\multirow{2}{*}{Reference} & liebe zuchauer guten abend .                                                                                                                                \\
                           & (love too much good evening .)                                                                                                                               \\ 
\hline
\multirow{2}{*}{Ours}      & liebe zuchauer guten abend .                                                                                                                                \\
                           & (love too much good evening .)                                                                                                                               \\ 
\hline
\multirow{2}{*}{Reference} & \begin{tabular}[c]{@{}l@{}}der wind weht meit mäßig im oten und an der ee richer wind\\mit tarken böen in den hochlagen der othälte turmböen .\end{tabular}  \\
                           & \begin{tabular}[c]{@{}l@{}}(The wind blows moderately in the east and at the east\\with tarke böen in the thaw of the eastern turbines .)\end{tabular}      \\ 
\hline
\multirow{2}{*}{Ours}      & \begin{tabular}[c]{@{}l@{}}der wind meist schwach mit mäßig bis mäßig mit\\schauern sonst kommt auch frisch sonst aus .\end{tabular}                         \\
                           & \begin{tabular}[c]{@{}l@{}}(The wind is usually weak with moderate to moderate\\with a good view, otherwise it will be fresh .)\end{tabular}                \\ 
\hline
\multirow{2}{*}{Reference} & \begin{tabular}[c]{@{}l@{}}und nun die wettervorherage ür\\morgen mittwoch den dreißigten märz .\end{tabular}                                                \\
                           & \begin{tabular}[c]{@{}l@{}}(and now the weather forecast \\for tomorrow, wednesday, march 30th .)\end{tabular}                                               \\ 
\hline
\multirow{2}{*}{Ours}      & \begin{tabular}[c]{@{}l@{}}und nun die wettervorhersage \\für morgen mittwoch den zwanzigsten oktober .\end{tabular}                                         \\
                           & \begin{tabular}[c]{@{}l@{}}(and now the weather forecast \\for tomorrow, wednesday, october 20th .)\end{tabular}                                             \\ 
\hline \hline
\multicolumn{2}{c}{KETI}                                                                                                                                                                \\ 
\hline
\multirow{2}{*}{Reference} & 가스가
  새고 있어요 .                                                                                                                                              \\
                           & (Gas is leaking .)                                                                                                                                           \\ 
\hline
\multirow{2}{*}{Ours}      & 가스가
  새고 있는 것 같아요 .                                                                                                                                          \\
                           & (I think gas is leaking .)                                                                                                                                   \\ 
\hline
\multirow{2}{*}{Reference} & 남편이
  높은데서 떨어져서 머리에 피나요 .                                                                                                                                    \\
                           & (My husband fell from a height and it bleeds
  from his head .)                                                                                              \\ 
\hline
\multirow{2}{*}{Ours}      & 남편이
  높은데서 떨어져서 머리에 피나요 .                                                                                                                                    \\
                           & (My husband fell from a height and it bleeds
  from his head .)                                                                                              \\ 
\hline
\multirow{2}{*}{Reference} & 술
  취한 사람이 방망이로 사람들을 때리고 있어요 .                                                                                                                               \\
                           & (A drunken man is hitting people with a bat .)                                                                                                               \\ 
\hline
\multirow{2}{*}{Ours}      & 술
  취한 사람이 방망이로 사람들을 때리고 있어요 .                                                                                                                               \\
                           & \begin{tabular}[c]{@{}l@{}}(A drunken man is hitting people with a bat .) \\\end{tabular}                                                                   
\end{tabular}%
}
\label{table11}
\end{table}
We compare the use of our normalization methods.
If our normalization method was not used, the standard normalization method was used.
And the number of all cases was experimented with by applying different frame selection methods.
Only skip sampling method was used for sampling, and only stochastic augmentation method was used for augmentation. 
Finally, the SPSS method, which is our method combining the two methods, is compared when applied.

Through this experiment, we prove that we perform better when using Customized Normalization, and also that we perform best when using our method, SPSS.

\section{Qualitative Results}
In this section we report our qualitative results.
When given a sign language video, we share the results generated by our SLT model.
In addition, when the KETI dataset is given, Korean and English expressions are shown in a table together.
(See the Table \ref{table11}).

PHOENIX predicts relatively short sentences well. 
However, when the sentence is longer, the end of the sentence is relatively difficult to predict.
KETI predicts almost all sentences well and translates almost accurately long sentences.

\section{Conclusion And Future Work}
In this paper, we proposed a keypoint-based SLT model and proposed an optimal sign video processing method suitable for it.
We introduced a new normalization method suitable for sign language video keypoint vector processing, which was not in previous studies.
In addition, we proposed a method of robustness over multiple datasets by setting the number of input frames appropriate for each dataset without setting the number of input frames with the same arbitrary fixed number $N$ for all datasets.
Finally, similar to the method of Camagoz et al\cite{thirteen}, a method of solving the SLT problem as an NMT problem was introduced.

To prove the excellence of this method, we experimented with both high-resolution datasets (KETI) of video frames and low-resolution datasets (RWTH-PHOENIX-Weather-2014 T) of video frames.
Through experimental evaluation, we have proved that our method is  robust in sign language datasets with various sizes.
In addition, our video processing method has the generality that can be used not only in the sign language translation task but also in many tasks that video processing.

In future work, we will implement Sign-to-Gloss-to-Text (S2G2T) based on our video processing method to achieve sign language translation using gloss.
Based on Camagoz \textit{et al}\cite{twentyeight} and Yin \textit{et al}\cite{twentyseven}, we expect better performance if we translate text after gloss translation.
In addition, it is expected that various NMT models will be applied to achieve better performance improvement than before.

\bibliographystyle{ieeetr}
\bibliography{ref}
\end{document}